\documentclass[review]{elsarticle}
\usepackage{framed,multirow}
\usepackage{amssymb}
\usepackage{tabularx}
\usepackage{xcolor}
\usepackage{graphicx}
\usepackage{subfig}
\usepackage{lineno,hyperref}
\modulolinenumbers[5]
\usepackage{float}
\usepackage{array}
\newcolumntype{M}[1]{>{\centering\arraybackslash}m{#1}}
\journal{Journal of \LaTeX\ Templates}

\biboptions{authoryear}

\begin{document}

\begin{frontmatter}

\title{Symbolic Semantic Segmentation and Interpretation of COVID-19 Lung Infections in Chest CT volumes based on Emergent Languages}

% Group authors per affiliation:
\author{Aritra Chowdhury\corref{cor1}}
\ead{aritra.chowdhury@ge.com}
\author{Alberto Santamaria-Pang}
\author{James R. Kubricht}
\author{Jianwei Qiu}
\author{Peter Tu}
\address{Artificial Intelligence, GE Research, 1 Research Circle, Niskayuna NY 12309}
%\fntext[myfootnote]{Since 1880}

% or include affiliations in footnotes:

\cortext[cor1]{Corresponding author}


\begin{abstract}
The coronavirus disease (COVID-19) has resulted in a pandemic crippling the a breadth of services critical to daily life. Segmentation of lung infections in computerized tomography (CT) slices could be be used to improve diagnosis and understanding of COVID-19 in patients. Deep learning has come a long way in providing tools to accurately characterize infections and lesions in CT scans. However, they lack interpretability because of their black box nature. Recent advances in methods addressing the grounding problem of artificial intelligence have resulted in techniques that can used to develop symbolic languages  to represent data in specific domains. Inspired by human communication of complex ideas through language, we propose a symbolic framework based on emergent languages for the segmentation of COVID-19 infections in CT scans of lungs. We model the cooperation between two artificial agents - a Sender and a Receiver. These agents synergistically cooperate using emergent symbolic language to solve the task of semantic segmentation. Our game theoretic approach is to model the cooperation between agents unlike adversarial models e.g. Generative Adversarial Networks (GANs). The Sender retrieves information from one of the higher layers of the deep network and generates a symbolic sentence   sampled from a categorical distribution of vocabularies. The Receiver ingests the stream of symbols and cogenerates the segmentation mask. A private emergent language is developed among the Sender and Receiver that forms the communication channel used to describe the task of segmentation of COVID infections. We augment existing state of the art semantic segmentation architectures with our symbolic generator to form symbolic segmentation models. Twenty-nine CT volumes from two different sources of lung infection data, resulting from COVID-19 are used in this work to demonstrate our approach. Our symbolic segmentation framework achieves state of the art performance for segmentation of lung infections caused by COVID-19.  Our results show direct interpretation of symbolic sentences to discriminate between normal and infected regions, infection morphology and image characteristics. We show state of the art results for segmentation of COVID-19 lung infections in CT.  Our approach is agnostic of the base segmentation model and can be used to augment any model to improve segmentation accuracy and interpretability. 
\end{abstract}


\end{frontmatter}

%\linenumbers

\section{Introduction}
The world has faced a major health crisis since December 2019, due to the novel coronavirus (COVID-19) (\cite{wang2020novel}),  also known as Sars-COV-2 (\cite{andersen2020proximal}). Over 6 million cases were reported  resulting in over 370,000 deaths  (\cite{dong2020interactive}) across 187 countries. A crisis of this scale and magnitude has yet to occur in modern civilization; the severity of future pandemics and the importance of efficient human response cannot be stressed enough. Large scale efforts have been initiated by global health organizations  and national governments for diagnosis, testing and potential cures for the virus (\cite{sheridan2020fast}). Reverse transcription polymerase chain reaction (RT-PCR) has been considered the gold standard for the screening of COVID-19. However, there is a severe lack of testing equipment for environments that prohibit accurate screening of suspected cases. In addition, the reliability of the RT-PCR test has been questioned due to the high number of false negatives (\cite{ai2020correlation}). This calls for taking a multi-modality approach for consistent and robust diagnosis of COVID-19 in patients. One approach is to complement the RT-PCR test with radiological techniques such as X-rays and CT scans (\cite{rubin2020role, shi2020review}). This will help to significantly reduce the false negative rate and provide doctors with an elaborate and multifacted understanding of the disease. Recent results have shown that chest CT analysis can be utilized to obtain high levels of predictive performance (\cite{ai2020correlation}).

\begin{figure}[H]
\centering
\hfill
\subfloat[Symbols: 189 663 277 277 925 103 155 155]{{\includegraphics[scale=.4]{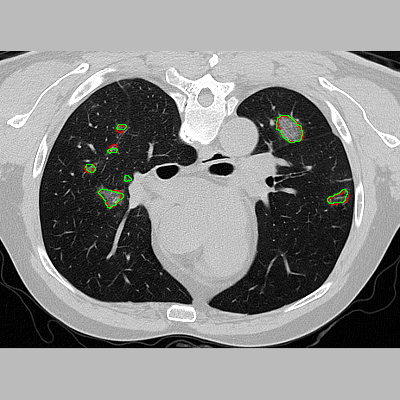} }}
\hfill
\subfloat[Symbols: 573 833 236 618 244 108 786 155]{{\includegraphics[scale=.4]{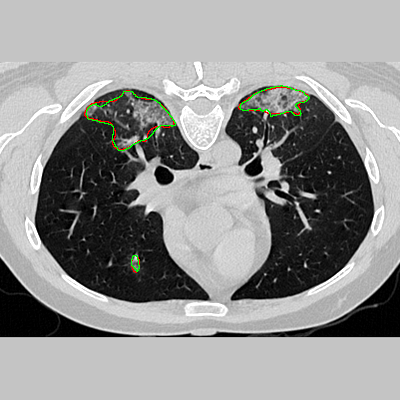} }}
\hfill
\caption{Examples of segmentation \textcolor{green}{ground truth} and \textcolor{red}{predictions} and the corresponding symbolic sentences on CT scan slices consisting of COVID-19 lung infections. We observe that our symbolic UNet provides accurate segmentation maps. In addition, the sentences provide clues towards interpreting the infections.}
\label{fig:Fig1}
\end{figure}

CT based analysis and diagnosis is generally preferred over X-rays because of access to three-dimensional views of organs (\cite{ye2020chest}). Typical signs of lung infections (e.g. ground-glass opacity) can be observed from CT slices  as shown in Fig. \ref{fig:Fig1}. The qualitative and quantitative appearance of the infection can provide important information related to detailed understanding of the characteristics of the COVID-19 disease. There are a number of challenges in segmentation of infections in chest CT slices because of the high variation in the size, texture and position of infections in the image. For example, small consolidations can result in false negative
detection outcomes. Deep learning based approaches to analysis of CT imagery has come a long way to address these issues (\cite{cheng2016computer}). However, as suspected, such inscrutable statistical models prove to be difficult to interpret. We propose a symbolic, game theoretic approach based on emergent languages to understand segmentation outputs in the context of lung infections in chest CT scans.
Current limitations in Artificial Intelligence (AI) include lack of interpretability and explainability; i.e. classical black-box approaches utilizing deep networks do not provide adequate evidence on how and why models perform the way they do (\cite{samek2017explainable}).  Explainability is considered to be of paramount importance in the medical field (\cite{london2019artificial}). This is necessary if we are to rely on AI and automated systems for clinical diagnosis and prognosis. In this work, we investigate synergies between deep learning based Semantic Segmentation (\cite{anthimopoulos2018semantic}) and Emergent Language (EL) (\cite {havrylov2017emergence}) models. We utilize properties of EL architectures to facilitate the interpretation of deep learning models and show how black box semantic segmentation can be extended to provide semantic sentences based on interpretable symbols. These sentences are sampled from a categorical distribution and subsequently integrated into state of the art segmentation architectures. We show, how we can significantly improve the performance of deep learning based segmentation networks by incorporating a symbolic layer that generates emergent language sentences.

In addition to the description and empirical analysis of the proposed methodology, we explore the utility of symbolic segmentation masks towards direct data interpretability in clinical applications. In this work, we utilize CT scans of patients afflicted with COVID-19 consisting of annotations of lung infections. We determine whether the symbolic sentences correspond to meaningful semantics in neural images. We show through rigorous experimentation, that symbolic segmentation networks are able to yield significant improvements over state of the art black box deep learning models. The symbols generated can also be used to interpret the results of the segmentation.

\label{sec1}

\section{Related work}
In this section, we detail relevant work in the area of segmentation of CT, medical image analysis of COVID-19 data, Emergent Languages and model interpretability in convolutional neural networks (CNNs)

\subsection{CT Segmentation}
CT imaging is an important modality for diagnosis of lung diseases like Pneumonia (\cite{sluimer2006computer}). Information obtained from high resolution CT data can provide important information to doctors for understanding diseases (\cite{gordaliza2018unsupervised}). Segmentation algorithms play a big part in accurately localizing nodules, lesions and infections in lungs. A lot of promising work has been done recently in the area of segmentation of chest CT data. An automated lung segmentation system based on bidirectional chain codes was presented in \cite{shen2015automated}. A number of deep learning approaches have been proposed as well to improve performance of segmentation in chest CT data. A central focussed CNN is proposed for segmentation of lung nodules in heterogenous CT (\cite{wang2017central}). GAN based synthetic data augmentation  was used to improve training of a discriminative model for lung segmentation in \cite{jin2018ct}.  A joint classification and segmentation model of an explainable COVID-19 system was proposed in \cite{wu2020jcs}. A semi supervised deep learning framework leveraging reverse and edge attention for segmentation of lung infections on COVID-19 was proposed in \cite{fan2020inf}.

\subsection{Medical Image Analysis of COVID-19}
Technologies leveraging artificial intelligence have been proposed to combat COVID-19 in multiple different ways at the patient scale (\cite{wang2020deep, chen2020deep}), the molecular scale (\cite{senior2020improved}) and societal scale (\cite{hu2020artificial}). Medical image analysis is usually applicable to analysing image data on the patient scale. A modification of the inception network was proposed in \cite{wang2020deep} for classifying COVID patients from normal controls. A UNet++ model was trained on 46,096 CT image slices from COVID patients in \cite{chen2020deep}. They show that the results of the model perform favorably when compared to expert radiologists' prediction. In addition, deep learning has also been used to segment infections in lung CT slices for downstream quantitative analysis for severity assessment (\cite{tang2020severity}), screening (\cite{shi2020large}) and lung infection quantification (\cite{rajinikanth2020harmony}) of COVID-19.

\subsection{Emergent Languages}
The emergent languages framework is inspired from \cite{lazaridou2016multi}, where the idea of using referential games for multi-agent cooperation is introduced. They show how the cooperative game leads to the emergence of an artificial language. These ideas are extended in \cite{havrylov2017emergence} by incorporating a sequence of symbols to further approximate sentence formation in emergent languages. The sequence of symbols is modeled using long short term memory networks (LSTMs). Introduction of natural language priors in models are also discussed here. Compositionality of emergent languages among multiple agents is discussed in \cite{cogswell2019emergence}.  A series of studies investigating the properties of protocols from the language is shown in \cite{lazaridou2018emergence}. Semantic action analysis using emergent languages is explored in \cite{santamaria2019towards}. The application of emergent languages to cell classification in pathology is explored in \cite{chowdhury2020escell}. Emergent languages has also been used to generate images using symbolic variational autoencoders (\cite{devaraj2020symbols}). An initial approach to symbolic segmentation was proposed recently in \cite{santamaria2020towards}.

\begin{figure}[H]
\centering
\includegraphics[width=\textwidth]{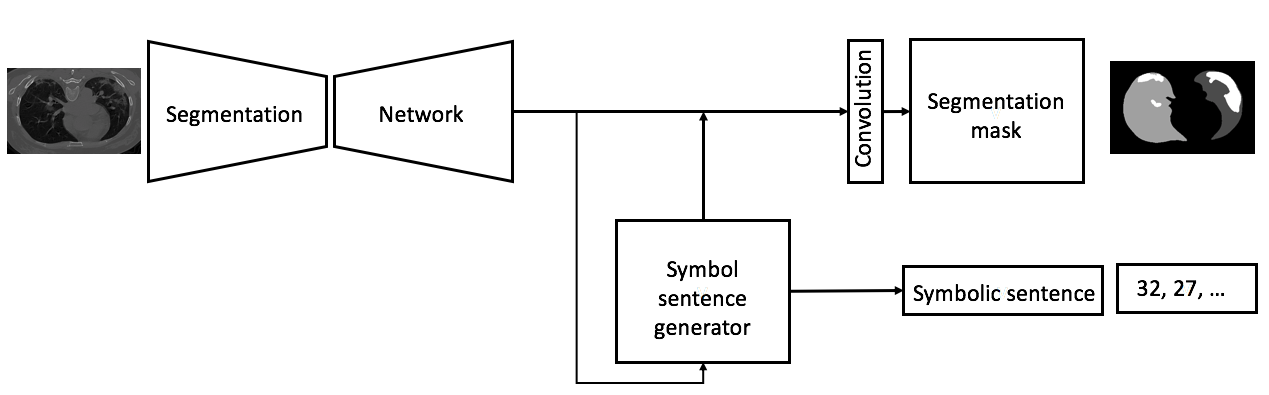}
\caption{Symbolic segmentation framework. Deep learning based segmentation networks are augmented using a symbolic generator that cogenerates sentences of the emergent language along with segmentation masks.}
\label{fig:sym_seg}
\end{figure}

\section{Methods}
We introduce the radical approach of symbolic semantic segmentation. Traditional semantic segmentation architectures like UNet are supplemented with a symbolic generator as shown in Fig. \ref{fig:sym_seg}.

We make the following assumptions to describe the methodology visualized in Figs. \ref{fig:sym_seg} and \ref{fig:SUNet},
\begin{enumerate}
\item There exists a segmentation network that provides a segmentation output $x$. 
\item There is a vocabulary $V = {w_{1}, w_{2}, ..., w_{N}}$, where $N$ is the size of the vocabulary. A sentence $S_{n}$ of length $n$ is a sequence of words or symbols ${w_{1}, w_{2}, ..., w_{n}}$.
\item A Sender agent or  network which receives the segmentation output $x$ and generates a sentence $S_{n}$ of length $n$, where $S_{n} = Sender(x)$.
\item A Receiver agent or network, which obtains the symbolic Sentence $S_{n}$ and generates an output $x' = Receiver(S_{n})$.
\item The final segmentation is co-generated from $x$ and $x'$.
\end{enumerate}

\subsection{Semantic segmentation}
In this work, we leverage three state of the art semantic segmentation architectures - UNet (\cite{ronneberger2015u}), UNet++ (\cite{zhou2018unet++}) and InfNet (\cite{fan2020inf}). UNet, introduced in 2015, was one of the first architectures to demonstrate how deep learning may be used to segment biomedical images. They demonstrated that the architecture was capable of fast and precise segmentation of neuronal structures in electron microscopic stacks. The UNet architecture consists of 3 sections - the contraction, the bottleneck and the expansion section. The contraction section consists of multiple contraction blocks made up of convolutional and pooling layers. The bottleneck layer, that mediates between the contraction and expansion sections, also consists of convolutional layers. The expansion section consists of multiple expansion blocks. These layers of convolutional and upsampling layers. Each expansion layer is appended by the corresponding feature maps in the contraction layers. This is what allows the architecture to preserve low level information required for accurately segmenting detailed images common in medical imaging. 
The UNet++ architecture, introduced in 2019 is an improvement over the UNet architecture. It uses the idea of Dense blocks from the DenseNet architecture (\cite{iandola2014densenet}) to improve performance. It differs from the original UNet in three ways. It consists of convolutional layers on skip pathways connecting the contraction and expansion layers. The skip connections have dense connections that improve gradient flow. They are also trained with dense supervision, that enables model pruning. The UNet++ architecture generates high resolution feature maps at multiple semantic levels. In addition the loss is estimated at four semantic levels.  The UNet++ model achieves significant performace gain over UNet.
InfNet is a segmentation network has been designed specifically for segmentation of lung infection caused by COVID-19 in CT scans. It consists of a parallel partial decoder that is used to aggregate a global feature map. Reverse attention and edge attention is used to model the boundaries to improve performance. They also introduce a semi supervised framework , COVID-SemiSeg to demonstrate state of the art performance on COVID CT data.

\subsection{Emergent Languages}

\begin{figure}[H]
\centering
\includegraphics[width=0.9\textwidth]{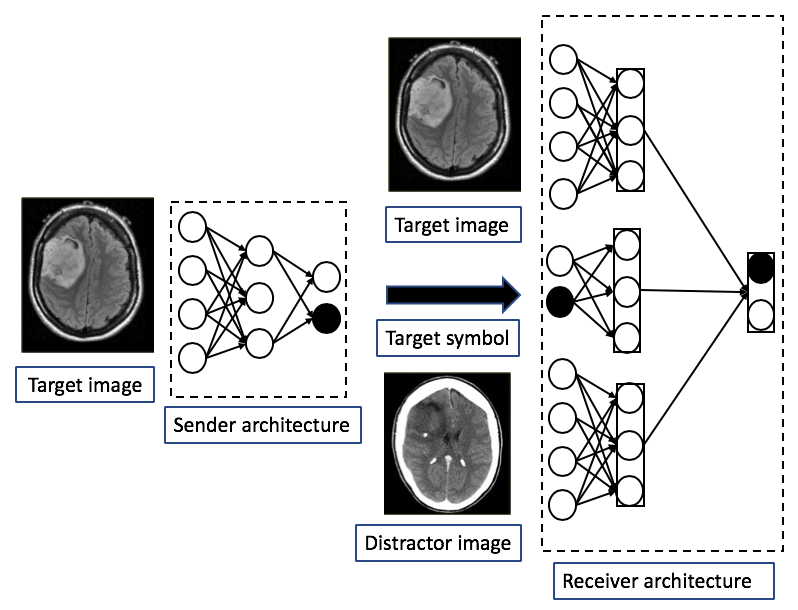}
\caption{Original emergent language framework. The Lewis signalling game involves a Sender and a Receiver. The Sender observes the target image and sends a symbol to the Receiver. The Receiver observes the target image, the Receiver and the symbol and the task of the Receiver is to pick out the target image correctly.}
\label{fig:EL}
\end{figure}

\begin{figure}[H]
\centering
\includegraphics[width=\textwidth]{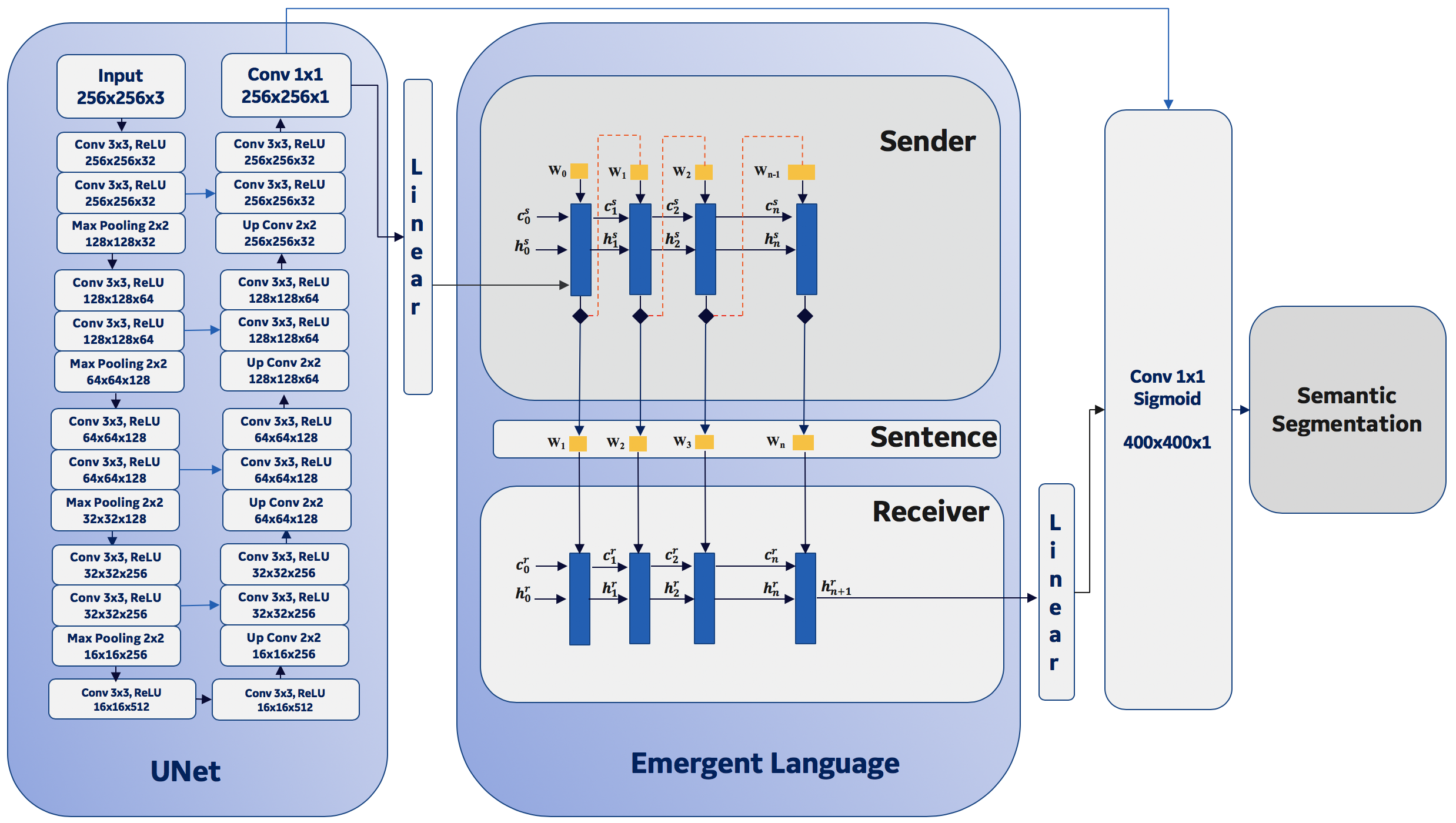}
\caption{SUNet segmentation architecture. The architecture consists of the baseline UNet model. The EL module takes as input the outut of the linear layer. The Receiver LSTM generates an output, that is concatenated with the output of the UNet and fed to a convolution layer and Sigmoid that generates the segmentation mask.}
\label{fig:SUNet}
\end{figure}

Fig. \ref{fig:EL} shows the original emergent language framework that was developed to solve the cooperative referential Lewis Signalling game (\cite{lazaridou2016multi}). 
The basic setup involves a sender architecture, a symbol generator and receiver architecture. The sender can be any network that extracts feature representations from input data. The sender sends the feature representations to the symbol generator where symbols are generated. These symbols are then fed to a receiver network that performs the classification. The only information that flows from the sender to the receiver are discrete representations instead of continuous features. In Fig. \ref{fig:EL}, the target image is an example of a CT scan of a brain with an indication. The sender generates a symbol using the symbolic generator. This symbol is then forwarded to the receiver network that observes the symbol, the target image and a distractor image (a normal CT scan of the brain). Using only the information in the symbol, the receiver must correctly guess distinguish the target image from the distractor image.
In this work, we	implemented a variant for Sender and Receiver networks as reported in \cite{havrylov2017emergence}, using stacked LSTM models (\cite{hochreiter1997lstm}).
The sender receiver emergent language module is shown in Fig. \ref{fig:SUNet}. The module in the middle consists of the Sender and Receiver LSTM models.

The input to the \textbf{Sender Network} is a tensor $x$ that can be the feature representation of the input image  $I$. A token $<S>$ represents the start of the message. The input is passed to a stacked LSTM network after performing a linear transformation. The initial hidden state and the cell state, represented as $h_{0}^{s}$ and $c_{0}^{s}$ are initialized to zero. The LSTM samples a single symbol from a categorical distribution $w~Cat(p_{v}^{n})$, where $p_{v}^{n}$ are the probabilities with respect to the symbols in the vocabulary $V$ at interation $n$. This operation is not differentiable and therefore gradients cannot be estimated for the backpropagation algorithm. The Gumble-Softmax (GS) trick )\cite{jang2016categorical}) is therefore used to relax the categorical distribution.  We estimate a symbol or word $w_{i}$ is sampled at each iteration $n$ according to Eq. \ref{eq:GS}.

\begin{equation}
w_{i} = G_{\tau}(p_{i}^{n}) = \frac{\exp(\log(p_{i}^{n}) + g_{i}) / \tau}{\sum_{j=1}^{v} \exp(\log(p_{j}^{n}) + g_{j}) / \tau}
\label{eq:GS}
\end{equation}
$\tau$ is the temperature parameter that regulates the GS operator $G_{\tau}$. The output of the sender is the final hidden state $h_{n+1}^{s}$ that encodes the sentence as a sequence of words $w_{i}$ as $h_{n+1}^s = LSTM(w_{i}, h_{n}^{s}, c_{n}^{s})$. At inference time, we do not apply the GS operator (\cite{jang2016categorical}) and normal categorical sampling is done, thus making $h_{n+1}^{s}$ fully deterministic. The generated sentence is represented as $S_{n} = Sender(x)$.

The \textbf{Receiver network} is implemented as a standard LSTM model unlike the Sender. The input to the Receiver is the final hidden state of the Sender that encodes the sentence $S_{n}$. We encode the catgorical variable as a one-hot vector during inference to generate a deterministic output. The initial hidden state $h_{0}^{r}$ and cell state $c_{0}^{r}$ are set to zero initially. A linear transformation $Linear(h_{n+1}^{r})$ is applied to the Receiver's last hidden state. 
The Sender and Receiver are encouraged to develop a communication protocol using the vocabulary provided to it in the form of sentences generated from the Sender LSTM. If the training is successful, which means that the optimization has converged, we conclude that a new emergent language has been produced. The output of the receiver is $x' = Receiver(S_{n})$.

\subsection{Symbolic semantic segmentation}
We present our Symbolic Semantic Segmentation framework for simultaneous generation of segmentation maps and emergent language. This is shown in Fig. \ref{fig:sym_seg}.

We demonstrate the symbolic framework using emergent languages on each of the segmentation architectures detailed above - UNet, UNet and InfNet. We denote their symbolic counterparts as Symbolic UNet (SUNet), Symbolic UNet++ (SUNet++) and Symbolic InfNet (SInfNet). For purposes of demonstration, we show the \textit{SUNet} architecture in \ref{fig:SUNet}. 
We omit the final \textit{Sigmoid} function in left of Fig. \ref{fig:SUNet} to generate an output $x$. The Emergent Language framework (middle) is used to generate another output $x'$. The output feature maps are combined by concatenation and applying the \textit{Sigmoid} function (right). The training of the entire symbolic neural network is done end-to-end using stochastic gradient descent for backpropagation. When the optimization converges, we conclude that an interpretable symbolic language has emrged. 
The architectures of SUNet++ and InfNet are identical except for the base architecture. Instead of UNet in Fig. \ref{fig:SUNet}, we replace with UNet++ and InfNet respectively.

\section{Experiments and Results}
We detail the experiments and results of our symbolic semantic segmentation framework. 

\subsection{Datasets}
\begin{figure}[H]
\centering
\subfloat[Example of data from \cite{jun1covid}]{{\includegraphics[scale=.35]{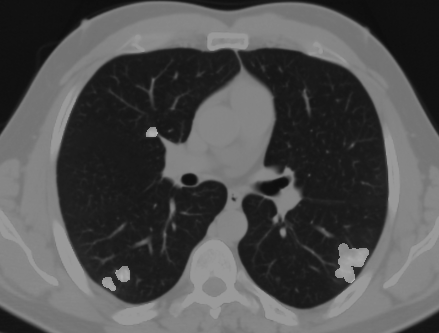} }}
\hfill
\subfloat[Example of data from \cite{radiopaedia}]{{\includegraphics[scale=.3]{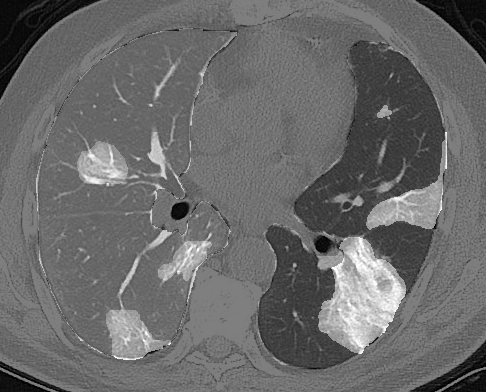} }}
\hfill
\caption{Example of CT data. The lung infections are shown as white overlays inside the lung in the CT slice. The CT data is gathered from two different sources. Preprocessing is performed in order to normalize the differences in appearance of the disparate data sources.}
\label{fig:examples_CT}
\end{figure}

In this work, we use volumetric CT scans from 2 different data sources - \cite{radiopaedia} and \cite{jun1covid}. We demonstrate our symbolic segmentation framework on 20 volumes from \cite{jun1covid} and 9 axial volumes from (\cite{radiopaedia}). The 9 volumes from \cite{radiopaedia} consist of both positive and negative COVID indications. The annotations have been created and segmented by a radiologist. An example of a slice from a positive scan is shown in Fig. \ref{fig:examples_CT} (right). The 20 volumes from \cite{jun1covid} consist of infections labelled by two radiologists and they have been verified by another experienced radiologist. They consist of segmentations of left lung, right lung and infections. However, in this work we only use the infection annotation. We use 26 volumes for training and 3 volumes for testing our results. 

\subsection{Pre-processing}
An example is shown in Fig. \ref{fig:examples_CT}(left). There are fundamental differences in the appearance of data from the two cohorts. For example, one of them is in 16 bit and the other is encoded in 8 bits. The intensity profile of each cohort is different as shown in Fig. \ref{fig:examples_CT}. Substantial preprocessing therefore needs to be done for our analysis.
A number of steps were applied prior to training. All the volumes contained a segmentation mask for lung and COVID-19 infected lung tissue. First, we cropped all volumes by having a distance of 20 voxels from the lung along the $x-y$ axis. Given that we have heterogenous datasets, we normalized all volumes according to  \cite{buda2019association} in the following manner: First, images were normalized to mean and standard deviation and standardized to have a maximum value of one. To account for images of different sizes, first we introduce zero padding to make images of same size in $x$ and $y$.  Then, every 2D slice was resized to 400x400 pixel units in the $x-y$ axis.

\subsection{Experimental setup}
We train a total of 6 different architectures. 3 are the baseline segmentation architectures - UNet, UNet++ and InfNet. The remaining 3 are their symbolic counterparts - SUNet, SUNet++ and SInfNet. The architectures of each of the Symbolic networks are constructed according to Figs. \ref{fig:sym_seg} and \ref{fig:SUNet}.

For each of the symbolic networks, we perform ablation experiments where we vary the the sentence length $N_{S} \in {8, 16}$  and vocabulary size $V \in {1000, 10000}$. We observe that the setting of $N_{S} = 8$ and $V=1000$ provide the best results. The results of this analysis is shown in Table \ref{table:ablation_expts}.
We use the default settings from each of the baseline architectures as described in the respective publications. The batch size for the experiments is set as $16$, the number of epochs for training is $300$, with early stopping on validation loss with a patience of $20$ epochs. The learning rate was set at $5e-5$ The images are resized to 400x400. The data is augmented using random rotation between -5 to +5 degrees and the scale is varied from  a factor 0.97 to 1.03. The Sender and Receiver embedding dimensions are set at 512.

\subsection{Results and discussions}
Table \ref{table:comparisons} shows the comparisons of segmentation metrics for each of the 6 architectures. We use Dice coefficient, Structure measure and Mean Absolute error (MAE) to measure the quality of segmentation (\cite{thoma2016survey}). The dice score and structure measure computes the amount of overlap between the prediction and the ground truth, so a higher number indicates a better segmentation. The mean absolute error measures the amount of dissimilarity between the output of the model and  the ground truth so a lower value is preferable.

\begin{center}
\begin{table}[h]
\centering
\begin{tabularx}{\textwidth} { 
  || >{\centering\arraybackslash}X 
  || >{\centering\arraybackslash}X 
  | >{\centering\arraybackslash}X
  | >{\centering\arraybackslash}X|| }
 \hline
  \hline
 Experiment & Dice score &  Structure measure & MAE\\ 
 \hline
 \hline
 UNet & 0.46 & 0.77 & 1.01 \\ 
 \hline
 SUNet & \textbf{0.71} & \textbf{0.83}  & \textbf{0.74}\\ 
 \hline
  UNet++ & 0.73 & 0.84 & 0.72\\ 
 \hline
   SUNet++ & \textbf{0.75} & \textbf{0.84} & \textbf{0.67}\\ 
 \hline
  InfNet & 0.75 & 0.85 & 0.71\\ 
 \hline
   SInfNet & \textbf{0.77} & \textbf{0.85} & \textbf{0.63}\\ 
 \hline
\end{tabularx}
\caption{Segmentation results comparison with baselines. The best performance is obtained using the Symbolic InfNet architecture (SInfNet) with a Dice score of 0.77. The symbolic versions of the architecture show significant improvement in performance over their baseline counterparts.}
\label{table:comparisons}
\end{table}
\end{center}
We observe from Table \ref{table:comparisons} that InfNet performs the best among baseline models. UNet++ does better that Unet which is expected. The important point to note is that each of the symbolic models perform better than their baseline counterparts, with the best performance overall being observed in SinfNet.

\begin{figure}[H]%
    \centering
    \subfloat{{\includegraphics[scale=.2]{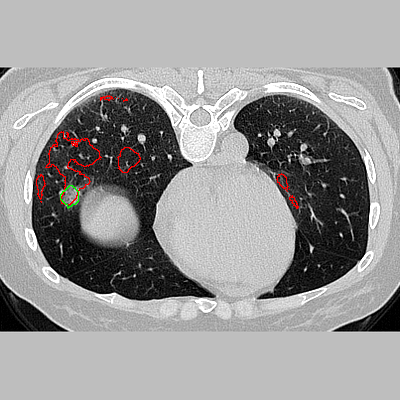} }}
	\hfill
    \subfloat{{\includegraphics[scale=.2]{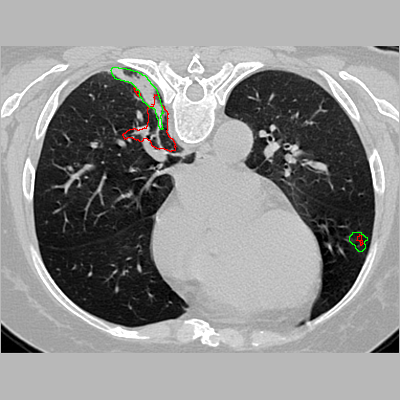} }}
	\hfill
    \subfloat{{\includegraphics[scale=.2]{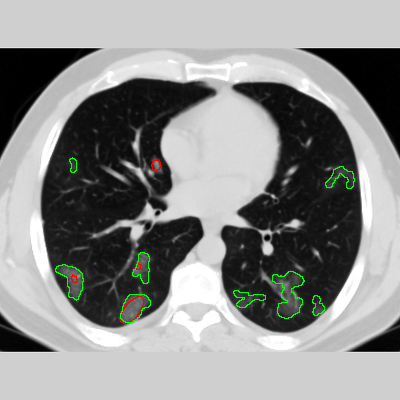} }}
    \hfill
    \subfloat{{\includegraphics[scale=.2]{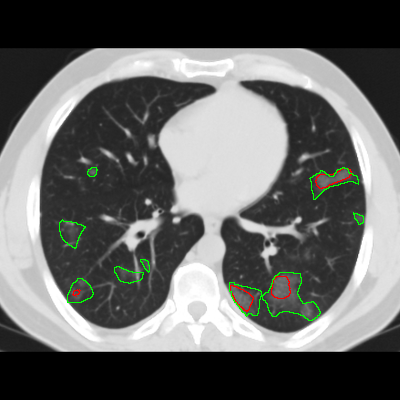} }}

	\subfloat{{\includegraphics[scale=.2]{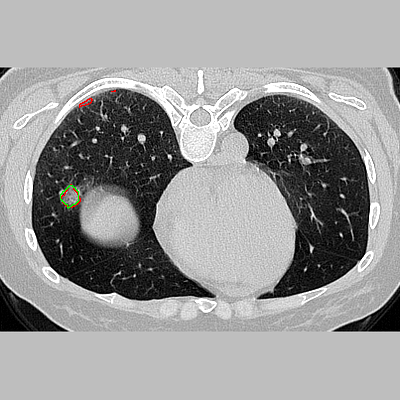} }}
	\hfill
    \subfloat{{\includegraphics[scale=.2]{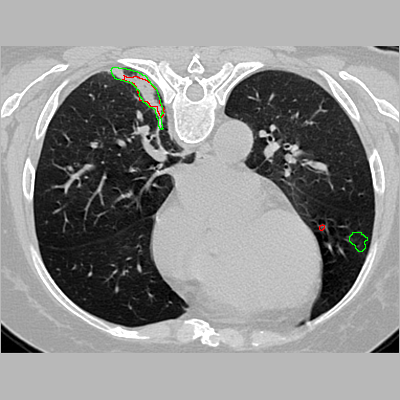} }}
	\hfill
    \subfloat{{\includegraphics[scale=.2]{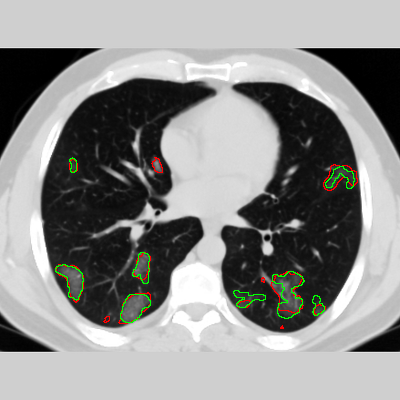} }}
    \hfill
    \subfloat{{\includegraphics[scale=.2]{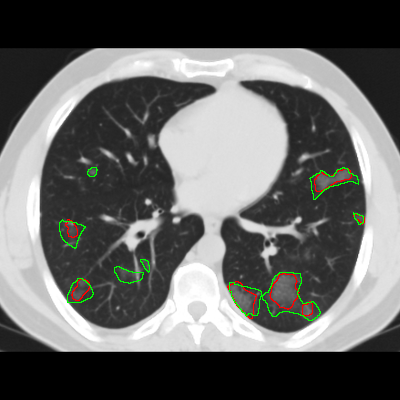} }}

	\subfloat{{\includegraphics[scale=.2]{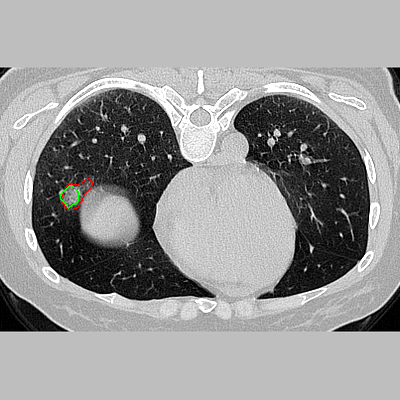} }}
	\hfill
    \subfloat{{\includegraphics[scale=.2]{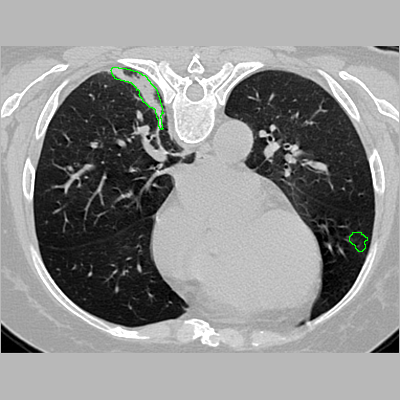} }}
	\hfill
    \subfloat{{\includegraphics[scale=.2]{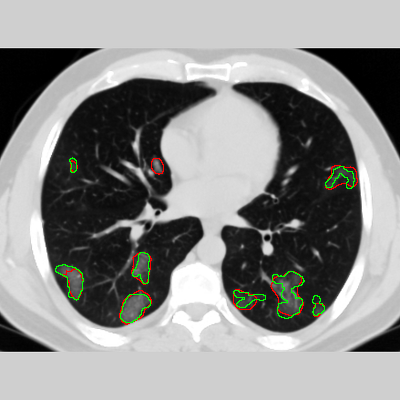} }}
    \hfill
    \subfloat{{\includegraphics[scale=.2]{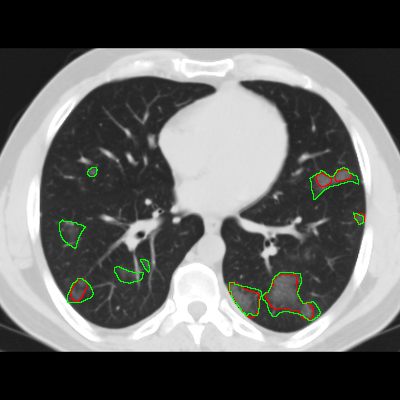} }}
\end{figure}

   \begin{figure}[H]%
    \centering
	\subfloat{{\includegraphics[scale=.2]{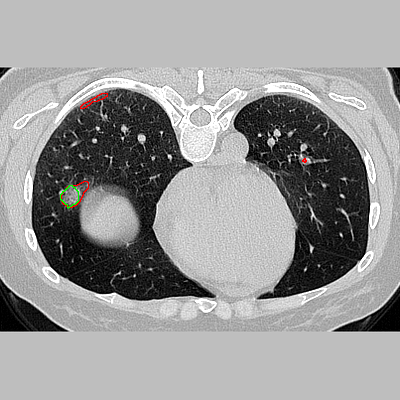} }}
	\hfill
    \subfloat{{\includegraphics[scale=.2]{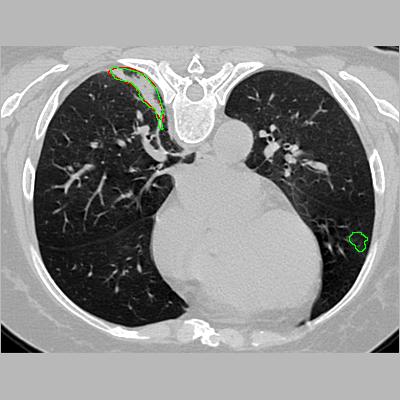} }}
	\hfill
    \subfloat{{\includegraphics[scale=.2]{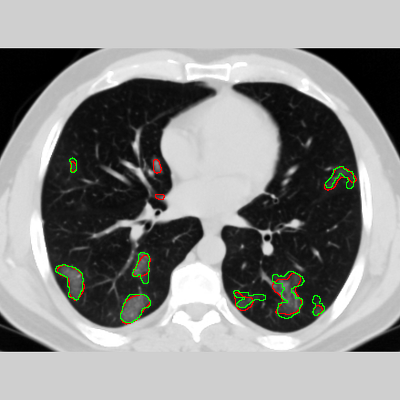} }}
    \hfill
    \subfloat{{\includegraphics[scale=.2]{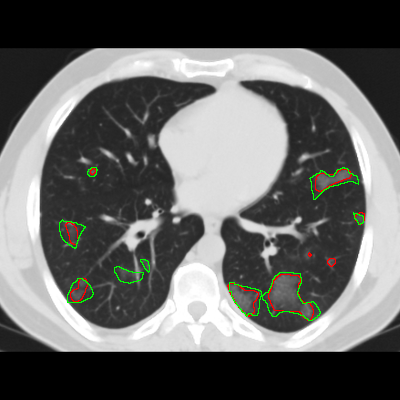} }}

    \subfloat{{\includegraphics[scale=.2]{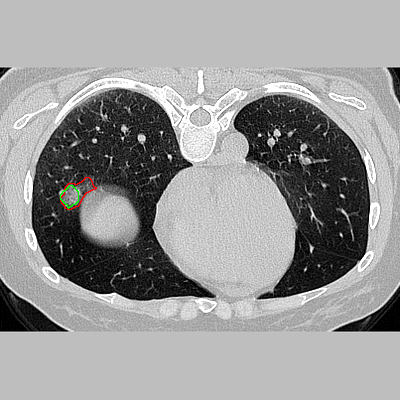} }}
	\hfill
    \subfloat{{\includegraphics[scale=.2]{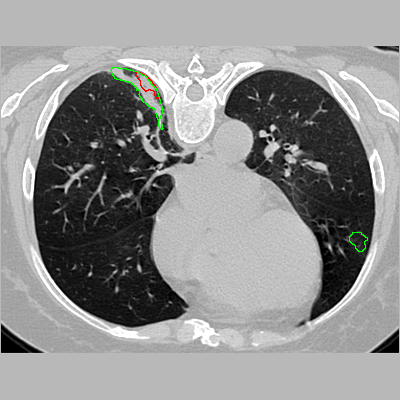} }}
	\hfill
    \subfloat{{\includegraphics[scale=.2]{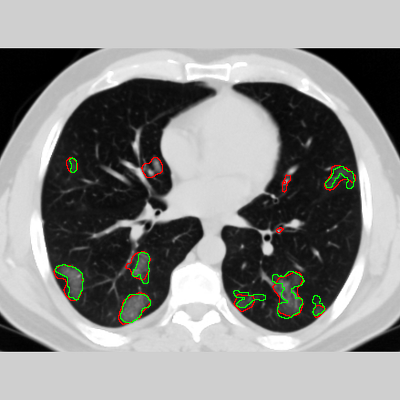} }}
    \hfill
    \subfloat{{\includegraphics[scale=.2]{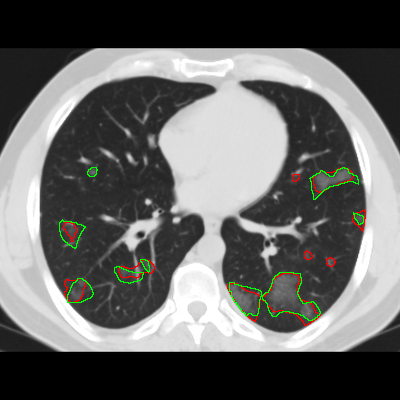} }}

   \subfloat{{\includegraphics[scale=.2]{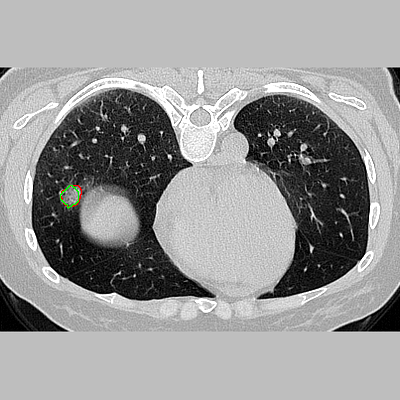} }}
	\hfill
    \subfloat{{\includegraphics[scale=.2]{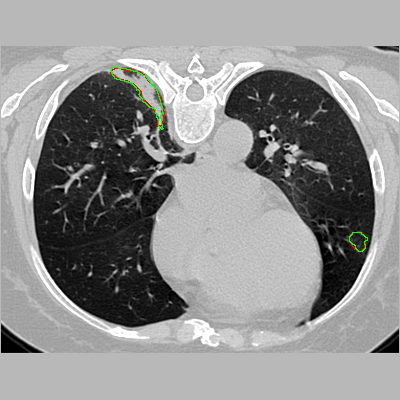} }}
	\hfill
    \subfloat{{\includegraphics[scale=.2]{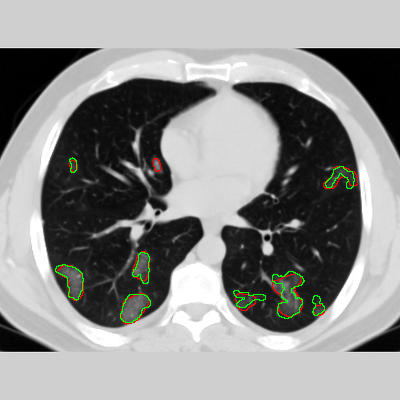} }}
    \hfill
    \subfloat{{\includegraphics[scale=.2]{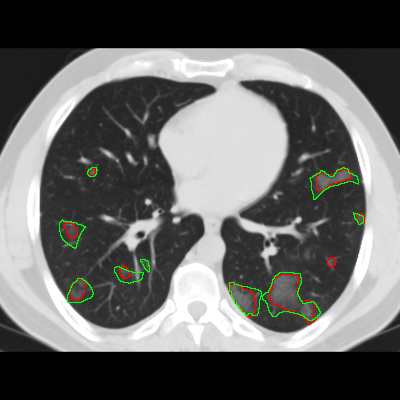} }}
    \caption{Segmentation results comparison with baselines. Each \textbf{column} represents a random CT slice with the \textcolor{green}{ground truth} and \textcolor{red}{prediction}. The \textbf{rows} are in the same order as \ref{table:comparisons}, \textbf{UNet(first)}, \textbf{SUNet(second)}, \textbf{UNet++(third)}, \textbf{SUNet++(fourth)}, \textbf{InfNet(fifth)}, \textbf{SInfNet(sixth)}. We observe that the quality of the predictions improve as we go from the first to the last row. We also observe that the outputs of the Symbolic network are significantly better than their baseline architectures}%
    \label{fig:Baselines}%
\end{figure}

Fig. \ref{fig:Baselines} visualizes the outputs for each of the 6 architectures in the same order as Table \ref{table:comparisons}. We qualitatively observe the same results as we found from the metrics in Table \ref{table:comparisons}. SinfNet shows the best quality of segmentation overall, and each of the symbolic networks perform significantly better than their baseline deep networks.

\begin{figure}[H]
\centering
\subfloat[Examples of from symbols of CT from cohort in \cite{jun1covid} ]{{\includegraphics[scale=.35]{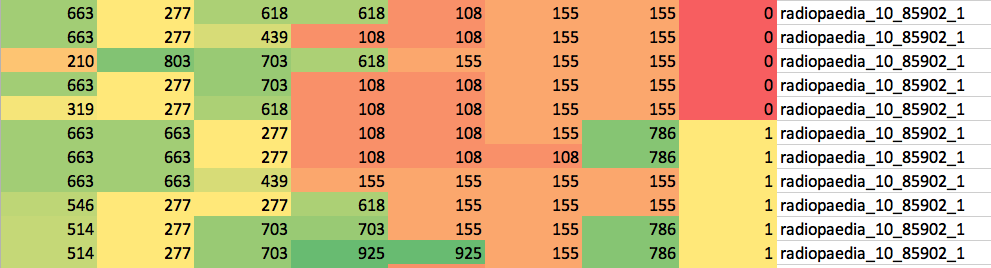} }} 

\subfloat[Examples of from symbols of CT from cohort in \cite{radiopaedia} ]{{\includegraphics[scale=.38]{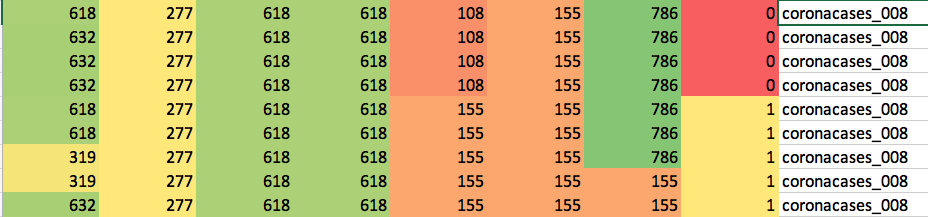} }}
\caption{Example of symbols for individual volumetric CT slices from the different cohorts. The \textbf{final column} represents the \textbf{cohort} and the \textbf{penultimate column} shows the \textbf{presence or absence of COVID-19}. The symbols are shown in the remaining columns. We oserve that the symbols are different for each CT slice. Similarity of the symbols indicate similarity in the features of input and outputs. Dissimilarity could denote a difference in the appearance of the infections or the input image.}
\label{fig:examples}
\end{figure}

Fig. \ref{fig:examples} shows the variation of the symbols generated from slices of chest CT with the presence and absence of COVID infections. We observe that the symbols seem to be different for every slice. Each symbol represents one or more phenotypic characteristics and features of the input image and the shape and appearance of infections in the output mask.
For example, the 5th symbol in Fig. \ref{fig:examples}(b) seems to correlate with the presence or absence of COVID-19 infection. 

\begin{figure}[H]
	\centering
	 \subfloat[595	428	475	497
]{{\includegraphics[scale=.2]{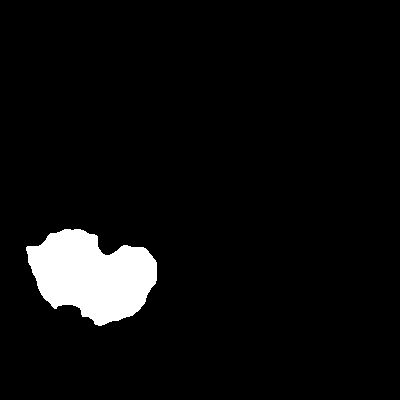} }}
	\hfill
    \subfloat[779	54	497	497
]{{\includegraphics[scale=.2]{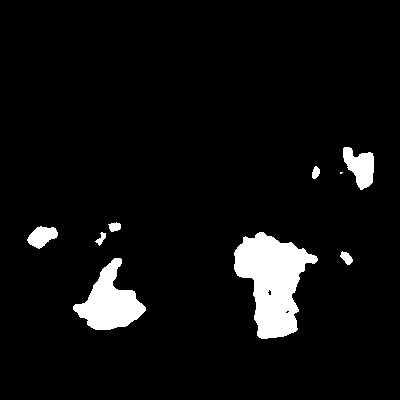} }}
	\hfill
    \subfloat[512	69	69	428
]{{\includegraphics[scale=.2]{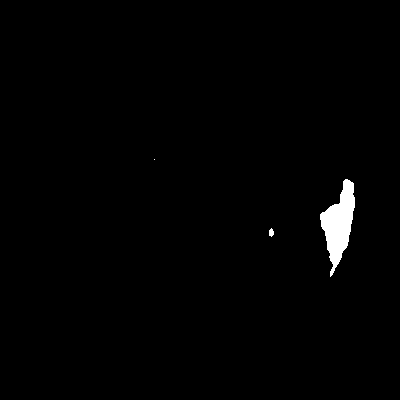} }}
    \hfill
    \subfloat[512	185	138	428
]{{\includegraphics[scale=.2]{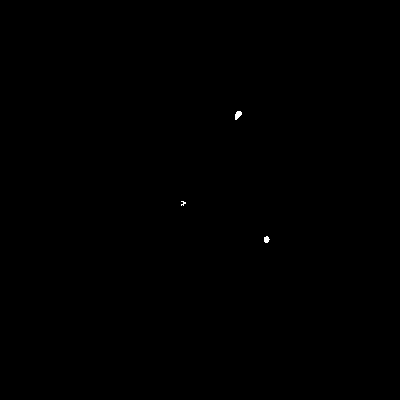} }}

    \subfloat[579	472	10	 670
]{{\includegraphics[scale=.2]{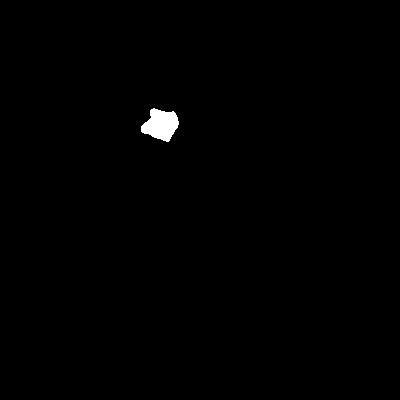} }}
	\hfill
    \subfloat[579	469	312	670
]{{\includegraphics[scale=.2]{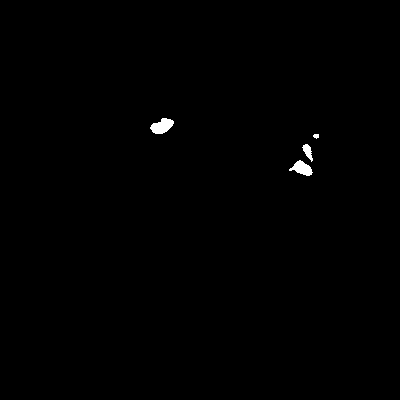} }}
	\hfill
    \subfloat[776	619	10	 622
]{{\includegraphics[scale=.2]{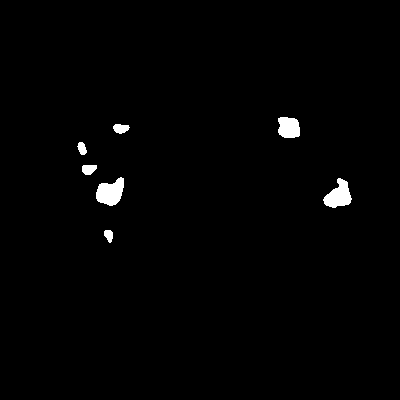} }}
    \hfill
    \subfloat[579	469	596	596
]{{\includegraphics[scale=.2]{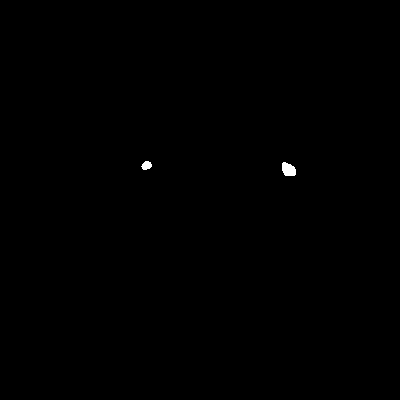} }}

    	\subfloat[573	618	618	618
]{{\includegraphics[scale=.2]{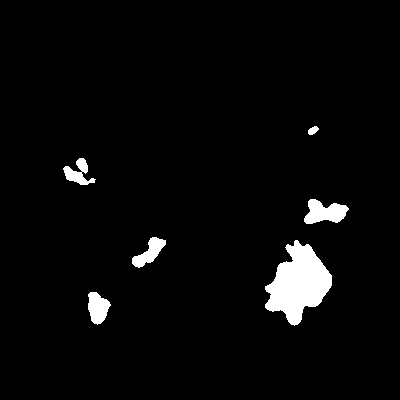} }}
	\hfill
    \subfloat[176	618	277	439
]{{\includegraphics[scale=.2]{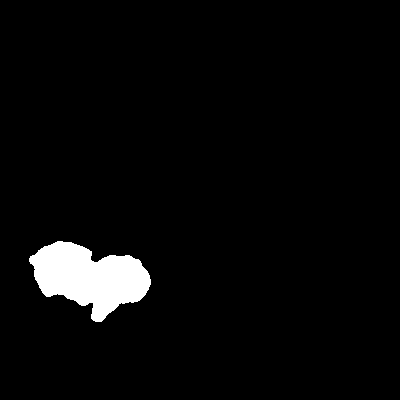} }}
	\hfill
    \subfloat[573	319	439	632
]{{\includegraphics[scale=.2]{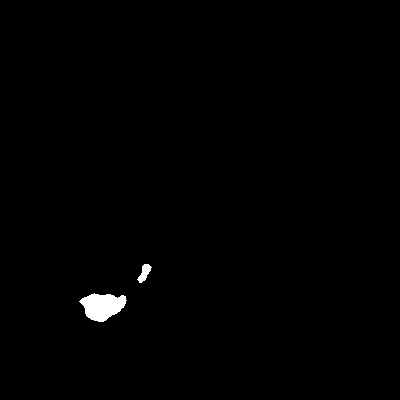} }}
    \hfill
    \subfloat[169	210	439	618
]{{\includegraphics[scale=.2]{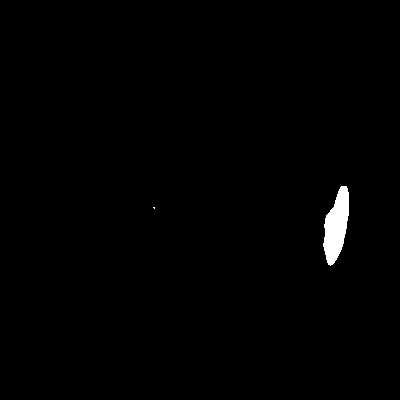} }}
\caption{Sampling of segmentation maps and the generated symbols. Here we only show the first 4 symbols out of 8 symbols, because they capture the most important features. Each row represents the 3 symbolic architectures. The \textbf{First row (SUNet)} shows different types of segmentation output maps. We observe that the symbols represent different types of output maps. The symbol \textbf{512} appears to represent small maps to the left of the image. The \textbf{Second row (SUNet++)} shows how the symbol 579 seems to represent smaller and scattered infections.  The \textbf{Third row (SInfNet)} shows different symbols for different types of infection scattering in the lungs.}
\label{fig:seg_symbols}
\end{figure}

Fig. \ref{fig:seg_symbols} shows segmentation outputs with the corresponding symbols. We observe that there appears to be semantically uniquely symbols or words that define a particular segmentation map. Each symbol embodies one or more semantically meaningful attributes of the masks. There appears to be certain symbols that correlate with the shape, size and locations of each of local areas of infection in the lung. For example, in Row 1 (SUNet outputs), the symbol 512 seems to correspond to small infection areas on the right lung. In Row 2 (SUNet++ outputs), the symbol 579 also appears on 3 of the segmentation maps. They could indicate small areas of infection. Row 3 corresponds to SInfNet outputs and the symbol 573 appears to be common between the first and third image.

An important consideration of all empirical work are ablation experiments. We show the performance of our symbolic semantic segmentation framework with respect to the important parameters of the emergent language, i.e. the length of the sentences ($N_{S}$) and the vocabulary size ($V$). Results of the ablation experiments are shown in the following table.

\begin{center}
\begin{table}[H]
\centering
\begin{tabular}{||m{5em}||m{3.5cm}|m{1.2cm}|m{2cm}|m{1cm}||}
 \hline
 \hline
 Experiment & Parameters & Dice score &  Structure measure & MAE\\ 
 \hline
 \hline
 \multirow{4}{4em}{SUNet} & $N_{S}=8, V=1000$ & \textbf{0.72} & \textbf{0.83} & \textbf{0.74} \\ 
 & $N_{S}=8,V=10000$ & 0.71 & 0.82 & 0.77 \\ 
 & $N_{S}=16,V=1000$ & 0.72 & 0.82 & 0.74 \\ 
 & $N_{S}=16,V=10000$ & 0.72 & 0.82 & 0.74 \\ 
 \hline
 \hline
 \multirow{4}{4em}{SUNet++} & $N_{S}=8, V=1000$ & \textbf{0.75} & \textbf{0.84} & \textbf{0.67} \\ 
 & $N_{S}=8,V=10000$ & 0.73 & 0.84 & 0.69 \\ 
 & $N_{S}=16,V=1000$ & 0.74 & 0.83 & 0.68 \\ 
 & $N_{S}=16,V=10000$ & 0.74 & 0.83 & 0.68 \\ 
 \hline
  \hline
 \multirow{4}{4em}{SInfNet} & $N_{S}=8, V=1000$ & \textbf{0.77} & \textbf{0.85} & \textbf{0.63} \\ 
 & $N_{S}=8,V=10000$ & 0.76 & 0.85 & 0.67 \\ 
 & $N_{S}=16,V=1000$ & 0.76 & 0.84 & 0.67 \\ 
 & $N_{S}=16,V=10000$ & 0.75 & 0.85 & 0.68 \\ 
 \hline
\end{tabular}
\caption{Ablation experiments. We show the ablation experiments by varying the sentence length of the symbols $N_{S}={8, 16}$ and vocabulary $V={1000, 10000}$.
We observe that the results are quite robust with respect to the parameters of the emergent language with a marginal performance improvement with the combination of $N_{S}=8$ and $V=1000$}.
\end{table}
\label{table:ablation_expts}
\end{center}

Table 2 shows the results of the ablation experiments. We observe here that the symbolic semantic framework is robust when we vary the crucial parameters of the emergent language layer, $N_{S}$ and $V$. In general for each of the 3 Symbolic models, we see that the combination $N_{S}=8$ and $V=1000$ appear to perform the best. We therefore use this combination when presenting the results in Table \ref{table:comparisons}. Also, this means that there is no additional information being added by increasing the sentence length and vocabulary size and $N_{S}=8$ and $V=1000$ are approximately optimal.

\begin{center}
                \begin{table}[H]
                                \centering
                                \begin{tabular}{||M{5em}||M{.8cm}M{.8cm}|M{1.3cm}M{1.3cm}|M{1.3cm}M{1.3cm}||}
                                                \hline
                                                \hline
                                                \multirow{2}{5em}{\centering Experiment} & \multicolumn{2}{c|}{Parameters} & \multicolumn{2}{c|}{COVID Presence} & \multicolumn{2}{c||}{COVID Area} \\
                                                & $N_S$ & $V$ & $S^*$ & $R^2_{\tiny\textrm{McFadden}}$ & $S^*$ & $r^2$ \\
                                                \hline
                                                \hline
                                                \multirow{4}{4em}{SUNet} & 8 & 1000 & $S_{3}$ & 0.21 & $S_{3}$ & 0.43 \\
                                                & 8 & 10000 & $\mathbf{S_{4}}$ & \textbf{0.43} & $\mathbf{S_{4}}$ & \textbf{0.63} \\
                                                & 16 & 1000 & $S_{4}$ & 0.28 & $S_{1}$ & 0.52 \\
                                                & 16 & 10000 & $S_{3}$ & 0.24 & $S_{1}$ & 0.43 \\
                                                \hline
                                                \multirow{4}{4em}{SUNet{\small++}} & 8 & 1000 & $S_{3}$ & 0.32 & $S_3$ & 0.42 \\
                                                & 8 & 10000 & $S_{2}$ & 0.25 & $S_{4}$ & 0.43 \\
                                                & 16 & 1000 & $S_{2}$ & 0.33 & $S_{2}$ & 0.66 \\
                                                & 16 & 10000 & $\mathbf{S_{3}}$ & \textbf{0.46} & $\mathbf{S_{3}}$ & \textbf{0.74} \\
                                                \hline
                                                \multirow{4}{4em}{SInfNet} & 8 & 1000 & $S_{2}$ & 0.19 & $S_{4}$ & 0.40 \\
                                                & 8 & 10000 & $\mathbf{S_{2}}$ & \textbf{0.53} & $S_{1}$ & 0.50 \\
                                                & 16 & 1000 & $S_{4}$ & 0.40 & $S_{3}$ & 0.48 \\
                                                & 16 & 10000 & $S_{1}$ & 0.52 & $\mathbf{S_{1}}$ & \textbf{0.66} \\
                                                \hline
\end{tabular}
 \caption{Results from logistic (COVID Presence) and linear (COVID Area) regression analyses using individual symbols as independent variables. Model performance is captured via McFadden's pseudo-$R^2$ for logistic regression (values between 0.2 and 0.4 indicate excellent fit) and squared Pearson correlation coefficient for linear regression. The $S^*$ column indicates which symbol in the sequence was most predictive of the corresponding measurement type.}
 \label{table:R2}
\end{table}
\end{center}
 
Results from Table \ref{fig:Baselines} and Fig. \ref{fig:Baselines} indicate that symbolic expressions can be used to successfully predict segmentation masks of lung infections in Chest CT data. Those symbols also appear to be informative according to the qualitative results depicted in Fig. \ref{fig:seg_symbols}. We performed further regression analyses to detemine whether individual symbols could predict the presence or absence of COVID and morphology (area) occupied by the infection. Specifically, we examined which expression symbol (i.e. first, second, etc.) is best at predicting the outcome of all candidate models. The results from the analysis are shown in Table \ref{table:R2}.

The statistical model used to predict each outcome varied. For binary data (presence or absence of COVID-19), a binary logistic regression model was used. The linear regression was performed on data where an infection was present. We report squared Pearson correlation coefficient $R^{2}$ (\cite{benesty2009pearson}) values for continuous outcomes and McFadden’s pseudo-$R^{2}$ (\cite{veall1994evaluating}) for categorical outcomes. Results indicate very high correlations between expression symbols and COVID presence and area, especially in models where a large vocabulary size was used. Optimal predictions for COVID presence were found using the second symbol in the expression (SInfNet model), whereas optimal predictions for area were found using the third symbol (SUNet++ model). All outcomes were best explained using a vocabulary size of 10000, although optimal sentence length seemed to vary between models. Taken together, the current results demonstrate that emergent language expressions generated in each of the proposed models carry a wealth of information about key concepts in medical imagery. 

\subsection{Limitations and future work}
Even though we introduce symbolic representations, the deep networks do not automatically become completely interpretable and transparent. However, our work is a first of it's kind towards combining the power of statistical deep learning with the  interpretable capacity of symbolic methods for medical imaging, particularly for segmentation of lung CT infections. It is no doubt that the sentences carry semantic information. We demonstrate preliminary methods of regression and qualitative analysis to try and interpret the meaning of the symbols. There are other sophisticated methods that maybe used to assign meaning and understand how the symbols interact with the input, output and with each other.

One avenue of future work, is to use saliency maps based notions of interpretability (\cite{selvaraju2017grad}). In essence, we would be able to map what each symbol represents with respect to regions in the input image. Another approach is to use the symbolic sentences in conjunction with natural language (\cite{lee2017emergent}), where we map the symbols and the vocabulary to a form of human understandable language like English.

\section{Conclusion}
The COVID-19 pandemic has brought the entire world to a standstill. We desperately need all the help we can to combat the disease. In order to fully understand and diagnose the disease, doctors use medical imaging modalities like CT to identify and characterize lung infections in possible COVID infected patients. Automated segmentation plays a big part in assisting radiologists to localize infections efficiently. In this work, we demonstrate how we can use symbolic semantic segmentation to segment lung infections with a high degree accuracy and interpretability. Our symbolic segmentation framework is built on top of LSTM based emergent languages. Using this framework, we are able to co-generate semantic segmentation maps and interpretable symbolic sentences. 

We show state-of-the art segmentation performance on CT data obtained from two cohorts. 
Moreover, we demonstrate the symbolic segmentation framework is flexible and can be used to augment any segmentation model to provide significant boost in performance.
The Symbolic InfNet (SInfNet) model that is built on top of the InfNet architecture achieves state-of-the-art Dice score of 0.77 on the validation data. We also show that each of the base models that we augment using the symbolic semantic segmentation framework (SUNet, SUNet++ and SInfNet) show significant increase in performance with respect to their baseline counterparts (UNet, UNet++ and InfNet respectively). These results are detailed in Table \ref{table:comparisons}.
 
Additionally, we show how the symbols maybe used as a tool for interpreting segmentation maps. Traditional deep learning systems are inherently blackbox in nature due to the continuous nature of their internal feature representations. The sentences generated from the segmentation can be used to analyse and query the model and quantify individual aspects and features of the segmentation masks. In Fig. \ref{fig:seg_symbols} we show how the symbols vary with respect to the appearance of the infections as observed on the segmentation masks. In addition, in Table \ref{table:R2}, we show how the symbols are correlated with phenotypes such as the area and presence of the COVID infection. Therefore, we consider our symbolic semantic segmentation framework to provide a different paradigm of deep learning based segmentation, where we use the emergent symbolic language to understand and interpret the models with respect to the inputs and outputs.

\bibliography{mybibfile}

\end{document}